\documentclass{article} 
\usepackage{nips11submit_e,times}
\usepackage{algorithm}
\usepackage{algorithmic}
\usepackage{graphicx}
\usepackage{tabularx}
\usepackage{amsfonts}
\usepackage{subfigure}
\usepackage{caption}

\usepackage{amssymb}
\usepackage{amsmath}
\usepackage{epstopdf}
\usepackage{verbatim}

\newcommand{\calM}{{\cal M}}
\newcommand{\calJ}{{\cal J}}
\newcommand{\calL}{{\cal L}}
\newcommand{\calP}{{\cal P}}
\newcommand{\calS}{{\cal S}}

\newcommand{\s}[1]{^{(#1)}}

\newcommand{\ssp}[4]{^{(#1),#2}_{#3_{#4}}}
\newcommand{\is}[4]{\pi^{(#1),#2}_{#3_#4}}
\newcommand{\st}[5]{a^{(#1),#2}_{#3_{#4},#3_{#5}}}

\newcommand{\Normal}[2]{{\cal N}(#1,#2)}
\newcommand{\argmax}{ \mathop{\mathrm{argmax}} }
\newcommand{\EV}{\mathrm{E}}

\newcommand{\refeqn}[1]{(\ref{#1})}

\newcommand{\eg}{e.g., }

\title{Tech Report \\ A Variational HEM Algorithm for Clustering\\ Hidden Markov Models}

\author{
Emanuele Coviello\thanks{ http://acsweb.ucsd.edu/$\sim$ecoviell/}\\
Department of Electrical and Computer Engineering\\
University of California, San Diego\\
9500 Gilman Drive
La Jolla, CA 92093\\
\texttt{ecoviell@ucsd.edu} \\
\AND
Anotni B. Chan\\
Department of Computer Science\\
City University of Hong Kong\\
Tat Chee Avenue,
Kowloon Tong, Hong Kong\\
\texttt{abchan@cityu.edu.hk} \\
\And 
 Gert R.G. Lanckriet\\
Department of Electrical and Computer Engineering\\
University of California, San Diego\\
9500 Gilman Drive
La Jolla, CA 92093\\
\texttt{gert@ece.ucsd.edu} \\
}

\nipsfinalcopy 
\begin{document}

\maketitle

\begin{abstract}
The hidden Markov model (HMM) is a generative model that treats sequential data under the assumption that each observation is conditioned on the state of a discrete hidden variable that evolves in time as a Markov chain.
In this paper, we derive a novel algorithm to cluster HMMs through their probability distributions.
We propose a hierarchical EM algorithm that i) clusters a given collection of HMMs into groups of HMMs that are similar, in terms of the distributions they represent, and ii) characterizes each group by a ``cluster center'', i.e., a novel HMM that is representative for the group.
We present several empirical studies that illustrate the benefits of the proposed algorithm.

\end{abstract}

\section{Introduction}

The hidden Markov model (HMM) \cite{Rabiner93} is a probabilistic model that assumes a signal is generated by a double embedded stochastic process. A hidden state process, which evolves over discrete time instants as a Markov chain, encodes the dynamics of the signal, and an observation process, at each time conditioned on the current state, encodes the appearance of the signal.
HMMs have been successfully applied to a variety of applications, including speech recognition~\cite{Rabiner93}, music analysis~\cite{qi2007music}, on-line hand-writing recognition~\cite{nag1986script}, analysis of biological sequences~\cite{krogh1994hidden}.


The focus of this paper is an algorithm for clustering HMMs. More precisely, we design an algorithm that, given a collection of HMMs, partitions them into $K$ clusters of ``similar'' HMMs, while also learning a representative HMM ``cluster center'' that summarizes each cluster appropriately. This is similar to standard k-means clustering, except that the data points are HMMs now instead of vectors in $\mathbb{R}^d$.
Such HMM clustering algorithm has various potential applications, ranging from
hierarchical clustering of sequential data (e.g., speech or motion sequences),
over hierarchical indexing for fast retrieval, to reducing the computational complexity of estimating mixtures of HMMs from large datasets via hierarchical modeling (e.g., to learn semantic annotation models for music and video). 

One possible approach is to group HMMs in \emph{parameter} space. However, as HMM parameters lie on a non-linear manifold, they cannot be clustered by a simple application of the k-means algorithm, which assumes real vectors in a Euclidean space. One solution, proposed in \cite{jebara2007}, first constructs an appropriate similarity matrix between all HMMs that are to be clustered (e.g., based on the Bhattacharya affinity, which depends on the HMM parameters \cite{jebara2004probability}),
and then applies spectral clustering. While this approach has proven successful to group HMMs into similar clusters~\cite{jebara2007}, it does not allow to generate novel HMMs as cluster centers. Instead, one is limited to representing each cluster by one of the given HMMs, e.g., the HMM which the spectral clustering procedure maps the closest to each spectral clustering center.
This may be suboptimal for various applications of HMM clustering.

An alternative to clustering the HMMs in parameter space is to cluster them \emph{directly} with respect to the \emph{probability distributions} they represent.
To cluster \emph{Gaussian} probability distributions, Vasconcelos and Lipmann \cite{Vasc1998} proposed a hierarchical expectation-maximization (HEM) algorithm. This algorithm starts from a Gaussian mixture model (GMM) with $K\s{b}$ components and reduces it to another GMM with fewer components, where each of the mixture components of the reduced GMM represents, i.e., \emph{clusters}, a group of the original Gaussian mixture components. More recently, Chan et al. \cite{Chan2010tr} derived an HEM algorithm to cluster  \emph{dynamic texture} (DT) models (i.e., linear dynamical systems, LDSs) through their probability distributions.
HEM has been applied successfully to construct GMM hierarchies for efficient image indexing~\cite{Vasc2001b},
to cluster video represented by DTs~\cite{Chan2010cvpr}, and to estimate GMMs or DT mixtures (DTMs,i.e., LDS mixtures) from large datasets for semantic annotation of images~\cite{carneiro07}, video~\cite{Chan2010cvpr} and music \cite{music:turnbull08,coviello2010}.


To extend the HEM framework for GMMs to mixtures of HMMs (H3Ms), additional marginalization of the hidden-state processes is required, as for DTMs. However, while Gaussians and DTs allow tractable inference in the E-step of HEM, this is no longer the case for HMMs. Therefore, in this work, we derive a variational formulation of the HEM algorithm (VHEM) to cluster HMMs through their probability distributions, based on a variational approximation derived by Hershey \cite{hershey2008variational}. The resulting algorithm not only allows to cluster HMMs, it also learns novel HMMs that are representative centers of each cluster, in a way that is consistent with the underlying generative probabilistic model of the HMM. The resulting VHEM algorithm can be generalized to handle other classes of graphical models, for which standard HEM would otherwise be intractable, by leveraging similar variational approximations. 
The efficacy of the VHEM-H3M algorithm is demonstrated for various applications, including hierarchical motion clustering, semantic music annotation, and online hand-writing recognition.

\section{The hidden Markov model}

A hidden Markov model (HMM) $\calM$ assumes a sequence of $\tau$ observations $y_{1:\tau}$ is generated by a double embedded stochastic process, where each observation $y_t$ at time $t$ depends on the state of a discrete hidden variable $x_t$ and where the sequence of hidden states $x_{1:\tau}$ evolves as a first order Markov process.
The discrete variables can take one of $N$ values, and the evolution of the hidden process is
encoded in a transition matrix $A = [a_{\beta,\gamma}]_{\beta,\gamma = 1,\dots,N}$ whose entries are the state transition probabilities  $a_{\beta,\gamma} = P(x_{t+1} = \gamma | x_t = \beta)$, and an initial state distribution $\pi = [\pi_1,\dots,\pi_N]$, where $\pi_\beta = P(x_1 = \beta)$.
Each state generates observation accordingly to an emission probability density function, $p(y_t|x_t = \beta, \calM )$, which here we assume to be a Gaussian mixture model:
\begin{eqnarray}
p(y| x =\beta) &=& \sum_{m = 1}^{M}  c_{\beta,m} \Normal{y;\mu_{\beta,m}}{\Sigma_{\beta,m}} 
\end{eqnarray}
where $M$ is the number of Gaussian components and $ c_{\beta,m}$ are the mixing weights.
In the following, when referring to a sequences of length $\tau$, we will use the notation $\pi_{x_{1:\tau}} P(x_{1:\tau} ) = \pi_{x_{1}} \prod_{t=2}^{\tau} a_{x_{t-1},x_{t}}$
to represent the probability that the HMM generates the state sequence  $x_{1:\tau}$.
The HMM is specified by the parameters $\calM = \{\pi, A, c_{\beta,m}, \mu_{\beta,m}, \Sigma_{\beta,m} \}$ which can be efficiently learned with the forward-backward algorithm \cite{Rabiner93}, which is based on maximum likelihood.


A hidden Markov mixture model (H3M) \cite{smyth1997} models a set observation sequences as samples from a group of $K$ hidden Markov models, which represent different sub-behaviors.
For a given sequence, an assignment variable $z \sim \mathrm{multinomial}(\omega_1, \cdots \omega_{K})$ selects the parameters of one of the $K$ HMMs, where the $k-th$ HMM is selected with probability $\omega_k$.
Each mixture component is parametrized by $\calM_z = \{\pi^z, A^z, c_{\beta,m}^z, \mu_{\beta,m}^z, \Sigma_{\beta,m}^z \}$ and the H3M is parametrized by $\calM = \{ \omega_z, \calM_z\}_{z=1}^{K}$.
Given a collection $\calS = \{ y_{1:\tau}^1,\dots,y_{1:\tau}^{|\calS|}\}$ of relevant observation sequences, the parameters of $\calM$ can be learned with recourse to the EM algorithm \cite{smyth1997}.

To reduce clutter, here we assume that all the HMMs have the same number of states $N$ and that all emission probabilities have $M$ mixture components, though a  more general case could be derived.

\section{Variational hierarchical EM algorithm for H3Ms}

The hierarchical expectation maximization algorithm (HEM)~\cite{Vasc1998} was initially proposed to cluster Gaussian distributions, by reducing a GMM with a large number of components to a new GMM with fewer components, and then extended to dynamic texture models~\cite{Chan2010tr}.
In this section we derive a variational formulation of the HEM algorithm (VHEM) to cluster HMMs.

\subsection{Formulation}


Let $\calM\s{b}$ be a base hidden Markov mixture model with $K\s{b}$ components.
The goal of the VHEM algorithm is to find a reduced mixture $\calM\s{r}$ with $K\s{r}<K\s{b}$ components that represent $\calM\s{b}$.
The likelihood of a random sequence $y_{1:\tau} \sim\calM\s{b}$ is given by 
	\begin{eqnarray}
	p(y_{1:\tau}|\calM\s{b}) = \sum_{i=1}^{K\s{b}} \omega_i\s{b} p(y_{1:\tau}|z\s{b}=i,\calM\s{b}) ,
	\label{eqn:mixmodel_base}
	\end{eqnarray}
 where $z\s{b}\sim\mathrm{multinomial}(\omega_1\s{b}, \cdots \omega_{K\s{b}}\s{b})$
 is the hidden variable that indexes the mixture components.  $p(y_{1:\tau}|z=i,\calM\s{b})$ is the likelihood of $y_{1:\tau}$ under the ith mixture component, and $\omega_i\s{b}$ is the prior weight for the ith component.  
The likelihood of the random sequence $y_{1:\tau}\sim\calM\s{r}$ is
	\begin{eqnarray}
	p(y_{1:\tau}|\calM\s{r}) = \sum_{j=1}^{K\s{r}} \omega_j\s{r} p(y_{1:\tau}|z\s{r}=j,\calM\s{r}) ,
	\label{eqn:mixmodel_reduce}
	\end{eqnarray}
where $z\s{r}\sim\mathrm{multinomial}(\omega_1\s{r},\cdots,\omega_{K\s{r}}\s{r})$ is the hidden variable for indexing components in $\calM\s{r}$.  Note that we will always use $i$ and $j$ to index the components of the base model, $\calM\s{b}$, and the reduced model, $\calM\s{r}$, respectively.  
In addition, we will always use $\beta$ and $\gamma$ to index the hidden states of $\calM\s{b}_i$ and $\rho$ and $\sigma$ for $\calM\s{r}_j$. 
To reduce clutter
 we will denote $p(y_{1:\tau}|z\s{b}=i,\calM\s{b}) = p(y_{1:\tau}|\calM\s{b}_i)$, and
$\EV_{y_{1:\tau}|\calM\s{b},z\s{b}=i} [\cdot ] = \EV_{\calM\s{b}_i} [\cdot ]$.
In addition, we will use short-hands $\calM\s{b}_{i,\beta_{1:\tau}}$ and $\calM\s{r}_{i,\rho_{1:\tau}}$ when conditioning over specific state sequences. For example, we denote $p(y_{1:\tau}| x_{1:\tau} = \beta_{1:\tau},\calM\s{b}_i) = p(y_{1:\tau}|\calM\s{b}_{i,\beta_{1:\tau}})$,
and $\EV_{y_{1:\tau}|\calM\s{b}_i,x_{1:\tau} = \beta_{1:\tau}} [\cdot ] = \EV_{\calM\s{b}_{i,\beta_{1:\tau}}} [\cdot ]$.
Finally, we will use $m$ and $\ell$ for indexing the gaussian mixture components of the emission probabilities of the base respectively reduced mixture, which we will denote as $\calM^{(b),i}_{\beta,m}$ and $\calM^{(r),j}_{\rho,\ell}$.

\subsection{Parameter estimation - a variational formulation}

To obtain the reduced model, we consider a set of $N$ virtual samples drawn from the base model $\calM\s{b}$, such that $N_i = N\omega\s{b}_i$ samples are drawn from the $i$th component.  We denote the set of $N_i$ virtual samples for the $i$th component as $Y_i = \{y\s{i,m}_{1:\tau}\}_{m=1}^{N_i}$, where $y\s{i,m}_{1:\tau}\sim \calM\s{b}_i$, and the entire set of $N$ samples as $Y = \{Y_i\}_{i=1}^{K\s{b}}$.  
Note that, in this formulation, we are not generating virtual samples $\{x\s{i,m}_{1:\tau}, y\s{i,m}_{1:\tau}\}$ according to the joint distribution of the base component, $p(x_{1:\tau},y_{1:\tau}|\calM\s{b}_i)$.  The reason is that the hidden state spaces of each base mixture component $\calM\s{b}_i$ may have a different representation, (\eg, the numbering of the hidden states may be permuted between the components).  This basis mismatch will cause problems when the parameters of $\calM\s{r}_j$ are computed from virtual samples of the hidden states of $\{\calM\s{b}_i\}$.  Instead, we must treat $X_i=\{x_{1:\tau}\s{i,m}\}$ as ``missing'' information, as in the standard EM formulation.

The likelihood of the virtual samples is
\begin{eqnarray}
	{\calJ({\calM\s{r}})} &=& \log p(Y|\calM\s{r}) = \sum_{i=1}^{K\s{b}} \log p(Y_i|\calM\s{r}) 
	\label{eqn:cost_function} = \sum_{i=1}^{K\s{b}} \log \sum_{j=1}^{K\s{r}} \omega\s{r}_j p(Y_i|\calM\s{r}_j).
\end{eqnarray}
In particular, for a given $\calM\s{r}$, the computation of $\log p(Y_i|\calM\s{r})$ can be carried out solving the optimization problems \cite{jordan1999introduction,Jaakkola00tutorialon}:
\begin{eqnarray}
	\log p(Y_i|\calM\s{r}) 
                                           &=& \max_{\calP_i(z_i)}\,  \log p(Y_i|\calM\s{r}) - D(\calP_i(z_i)|| P(z_i = j | Y_i, \calM\s{r})) \\
 &=&\max_{\calP_i(z_i)}\, \sum_j \calP_i(z_i = j) \left[  \log \omega\s{r}_j + \log p(Y_i|\calM\s{r}_j) - \log \calP_i(z_i = j) \right]
	\label{eqn:cost_one_sample}
\end{eqnarray}
for $i = 1,\,\dots,\,K\s{b}$,
where $\calP_i(z_i)$ are variational distributions and $D(p\|q) = \int p(y) \log \frac{p(y)}{q(y)} dy$ is the Kullback-Leibler (KL) divergence between two distributions, $p$ and $q$.
In order to obtain a consistent clustering \cite{Vasc1998}, we assume the whole sample $Y_i$ is assigned to the same component of the reduced model, i.e., $ \calP_i(z_i = j) = z_{ij} $, with $\sum_{j = 1}^{K\s{r}} z_{ij} = 1,\, \forall i$ and $z_{ij} \geq 0\, \forall i,j$, and \refeqn{eqn:cost_one_sample} becomes:
\begin{eqnarray}
\log p(Y_i|\calM\s{r}) 
 &=&\max_{z_{ij}}\, \sum_j z_{ij} \left[  \log \omega\s{r}_j + \log p(Y_i|\calM\s{r}_j) - \log z_{ij}  \right]
	\label{eqn:cost_one_sample_z}
\end{eqnarray}

Considering that virtual samples $Y_i$ are independent for different values of $i$,  
we can solve \refeqn{eqn:cost_one_sample} independently for each $i$, using the result in Section \ref{sec:opt2}, and find
\begin{eqnarray}
	\hat{z}_{ij} = \frac{\omega\s{r}_j \exp\{N_i \EV_{\calM\s{b}_i} \left[\log p(Y_i|{\calM}\s{r}_j)\right] \}}{ \sum_{j'=1}^{K\s{r}}\omega\s{r}_{j'} \exp\{N_i \EV_{\calM\s{b}_i} \left[\log p(Y_i|{\calM}\s{r}_{j'})\right] \}}.
	\label{eqn:zij_old}
\end{eqnarray}
For the likelihood of the virtual samples $p(Y_i|{\calM}\s{r}_j)$ we use
	\begin{eqnarray}
	\log p(Y_i|{\calM}\s{r}_j) &=& \sum_{m=1}^{N_i} \log p(y\s{i,m}_{1:\tau}|\calM\s{r}_j)
	= N_i \left[ \frac{1}{N_i} \sum_{m=1}^{N_i} \log p(y\s{i,m}_{1:\tau}|\calM\s{r}_j)\right] \\
	&\approx& N_i \EV_{\calM\s{b}_i} \left[\log p(y_{1:\tau}|\calM\s{r}_j)\right]
	\label{eqn:ExpLog}
	\end{eqnarray}
 where \refeqn{eqn:ExpLog} follows from the law of large numbers \cite{Vasc1998} (as $N_i \rightarrow \infty$).
Substituting \refeqn{eqn:ExpLog} in \refeqn{eqn:zij_old} we get the formula  for $z_{ij}$ derived in \cite{Vasc1998}:
\begin{eqnarray}
	\hat{z}_{ij} = \frac{\omega\s{r}_j \exp\{N_i \EV_{\calM\s{b}_i} \left[\log p(y_{1:\tau}|\calM\s{r}_j)\right] \}}{ \sum_{j'=1}^{K\s{r}}\omega\s{r}_{j'} \exp\{N_i \EV_{\calM\s{b}_i} \left[\log p(y_{1:\tau}|\calM\s{r}_{j'})\right] \}}.
	\label{eqn:zij}
\end{eqnarray}

We follow a similar approach to compute the expected log-likelihoods $\EV_{\calM\s{b}_i} \left[\log p(y_{1:\tau}|\calM\s{r}_j)\right]$.
We introduce variational distributions $\calP^{i,j}_{\beta_{1:\tau}}$ to approximate  $P(x_{1:\tau}  | y_{1:\tau}  , \calM\s{r}_{j} )$ for observations $y_{1:\tau} \sim \calM\s{b}_{i}$ emitted by state sequence $\beta_{1:\tau} $, and solve the maximization problem
\begin{eqnarray}
&\EV& \!\!\!\!\!\!{}_{\calM\s{b}_i} \left[\log p(y_{1:\tau}|\calM\s{r}_j)\right]  =\\
 &=& \max_{\boldsymbol \calP^{i,j}} \sum_{\beta_{1:\tau} } \pi\ssp{b}{i}{x}{1:\tau} \EV_{ \calM\s{b}_{i,\beta_{1:\tau}}} \left[\log p(y_{1:\tau} |\calM\s{r}_j)- D(\calP^{i,j}_{\beta_{1:\tau} } || P(x_{1:\tau}  | y_{1:\tau}  , \calM\s{r}_j ))\right] \label{eqn:exp_log_kl_tot}\\ 
&=& \max_{\boldsymbol \calP^{i,j}} \sum_{\beta_{1:\tau}} \pi\ssp{b}{i}{x}{1:\tau} \EV_{ \calM\s{b}_{i,\beta_{1:\tau}}} \left[ \sum_{\rho _{1:\tau}} \calP^{i,j}_{\beta_{1:\tau}}(x_{1:\tau}=\rho_{1:\tau}) \log \frac{\pi^{(r),j}_{\rho_{1:\tau}}p(y_{1:\tau} | \calM\s{r}_{j,\rho_{1:\tau}} )}{\calP^{i,j}_{\beta_{1:\tau}}(x_{1:\tau} = \rho_{1:\tau})}\right] 
\label{eqn:exp_log_kl}
\end{eqnarray}
where in \refeqn{eqn:exp_log_kl_tot} we have used the law of total probability to condition the expectation over each  state sequence $\beta_{1:\tau}$ of $\calM\s{b}_{i} $

In general, maximizing \refeqn{eqn:exp_log_kl} exactly sets the variational distribution to the true posterior and reduces (together with \refeqn{eqn:zij}) to the E-step of the HEM algorithm for hidden state models derived in \cite{Chan2010cvpr}:
\begin{eqnarray}
	 \EV_{\calM\s{b}_i} \left[\log p(y_{1:\tau}|\calM\s{r}_j)\right] &=& \EV_{\calM\s{b}_i} \left[  \EV_{x_{1:\tau} | \hat\calM\s{r}_j} \left[  \log p(x_{1:\tau},y_{1:\tau}|\calM\s{r}_j)   \right] \right] + \tilde{ {\cal H}}
\label{eqn:2exps}
\end{eqnarray}
where the inner expectation is taken with respect to the current estimate $\hat\calM\s{r}_j$ of $\calM\s{r}_j$, and $\tilde{ {\cal H}}$ is some term that does not depend on $\calM\s{r}_j$.

\subsection{The variational HEM for HMMs}

The maximization of \refeqn{eqn:exp_log_kl} cannot be carried out in a efficient way, as it involves computing the expected log-likelihood of a mixture. To  make it tractable we follow a variational approximation proposed by Hershey  \cite{hershey2008variational}, and restrict the maximization to factored distribution in the form of a Markov chain, i.e., 
\begin{eqnarray}
\calP^{i,j}_{\beta_{1:\tau}}(x_{1:\tau}=\rho_{1:\tau}) = \phi^{i,j}_{\rho_{1:\tau}|\beta_{1:\tau}}  = \phi^{i,j}_1(\rho_1,\beta_1) \prod_{t=2}^{\tau} \phi^{i,j}_t(\rho_{t-1},\rho_t,\beta_t)
\label{eg:hersheyHMM}
\end{eqnarray}
 where $\sum_{\rho = 1}^{N}  \phi^{i,j}_1(\rho_1,\beta_1) = 1\,\, \forall \beta_1$ and $\sum_{\rho = 1}^{N}   \phi^{i,j}_t(\rho_{t-1},\rho_t,\beta_t)=1 \,\, \forall \beta_t,\rho_{t-1}$.

Substituting \refeqn{eg:hersheyHMM} into \refeqn {eqn:exp_log_kl} we get a lower bound to the expected log-likelihood \refeqn{eqn:exp_log_kl}, i.e.,
\begin{eqnarray}
 \EV_{\calM\s{b}_i} \left[\log p(y_{1:\tau}|\calM\s{r}_j)\right] &\geq& 
  \calJ^{i,j} (\calM\s{r}_j, \boldsymbol\phi^{i,j}) \quad \forall \boldsymbol{\phi}^{i,j}
\end{eqnarray}
where we have defined
\begin{eqnarray}
 \calJ^{i,j} (\calM\s{r}_j,\boldsymbol\phi^{i,j}) = \sum_{\beta_{1:\tau}} \pi\ssp{b}{i}{\beta}{1:\tau}
\sum_{\rho_{1:\tau}} \phi^{i,j}_{\rho_{1:\tau}|\beta_{1:\tau}} \log \frac{\pi\ssp{r}{j}{\rho}{1:\tau} \exp \EV_{\calM\s{b}_{i,\beta_{1:\tau}}} [\log p(y_{1:\tau} | \calM\s{r}_{j,\rho_{1:\tau},} )]}{\phi^{i,j}_{\rho_{1:\tau}|\beta_{1:\tau}}}.
\label{eqn:exp_log_hmm}
\end{eqnarray}

Using the property of HMM with memory one that observations at different time instants are independent given the corresponding states, we can break the expectation term in equation \refeqn{eqn:exp_log_hmm} in the following summation
\begin{eqnarray}
	 \EV_{\calM\s{b}_{i,\beta_{1:\tau}}} [\log p(y_{1:\tau} | \calM\s{r}_{j,\rho_{1:\tau}} )] 
          = \sum_{t=1}^{\tau}  L(\calM\s{b}_{i,\beta_t} || \calM\s{r}_{j,\rho_t})
\label{eqn:exp_ll_1seq}
\end{eqnarray}
where $ L(\calM\s{b}_{i,\beta_t} || \calM\s{r}_{j,\rho_t}) =\EV_{ \calM\s{b}_{i,\beta_t}}  [\log p(y_t| \calM\s{r}_{j,\rho_t} )]  $.
As the emission probabilities are GMMs, the computation \refeqn{eqn:exp_ll_1seq} cannot be carried out efficiently.
Hence, we use a variational approximation \cite{hershey2007approximating}, and introducte variational parameters $\eta^{(i,\beta),(j,\rho)}_{\ell|m}$ for $\ell,m = 1,\dots,M$,
with $\sum_{\ell=1}^{M} \eta^{(i,\beta),(j,\rho)}_{\ell|m} = 1 \, \forall m$, and $\eta^{(i,\beta),(j,\rho)}_{\ell|m} \geq 0 \, \forall \ell,\!m$.
Intuitively,  $\boldsymbol\eta^{(i,\beta),(j,\rho)}$ is the responsibility matrix between gaussian observation components for state $\beta$ in $\calM\s{b}_i$ and state $\rho$ in $\calM\s{r}_j$, where $\eta^{(i,\beta),(j,\rho)}_{\ell|m}$ means the probability that an observation from component $m$ of $\calM\s{b}_{i,\beta}$ corresponds to component $\ell$ of $\calM\s{r}_{j,\rho}$. 
Again, we obtain a lower bound:
\begin{eqnarray}
	 L(\calM\s{b}_{i,\beta_t} || \calM\s{r}_{j,\rho_t}) &\geq& 
	 \calL(\calM\s{b}_{i,\beta_t} || \calM\s{r}_{j,\rho_t}) \quad \forall \boldsymbol\eta^{(i,\beta),(j,\rho)}
\end{eqnarray}
where we have defined:
\begin{eqnarray}
	\calL(\calM\s{b}_{i,\beta_t} || \calM\s{r}_{j,\rho_t})  = \sum_{m=1}^{M} c^{(b),i}_{\beta,m}   \sum_{\ell=1}^{M}  \eta^{(i,\beta),(j,\rho)}_{\ell|m} \left[ \log c^{(r),j}_{\rho,\ell} + L_{G}(\calM^{(b),i}_{\beta,m}|| \calM^{(r),j}_{\rho,\ell}) - \log \eta^{(i,\beta),(j,\rho)}_{\ell|m} \right]
\label{eqn:exp_ll_gmms}
\end{eqnarray}
where $L_{G}(\calM^{(b),i}_{\beta,m}|| \calM^{(r),j}_{\rho,\ell}) = \EV_{y | \calM^{(b),i}_{\beta,m}} [\log P(y |\calM^{(r),j}_{\rho,\ell})] $ can be computed exactly for Gaussians
\begin{eqnarray}
        L_{G}(\calM^{(b),i}_{\beta,m}|| \calM^{(r),j}_{\rho,\ell})
&=& -\frac{1}{2} d\log 2\pi + \log \left|\Sigma\s{r}_{j,\rho}\right|  -\frac{1}{2} \mbox{tr} \left({\Sigma\s{r}_{j,\rho}}^{-1} \Sigma\s{b}_{i,\beta}\right)    \\ 
 && - \frac{1}{2} (\mu\s{r}_{j,\rho} - \mu\s{b}_{i,\beta})^T{\Sigma\s{r}_{j,\rho}}^{-1} (\mu\s{r}_{j,\rho} - \mu\s{b}_{i,\beta}) .
        \label{eqn:exp_log_g}
\end{eqnarray}

Plugging \refeqn{eqn:exp_ll_gmms} into \refeqn{eqn:exp_ll_1seq} and \refeqn{eqn:exp_log_hmm} we get the lower bound to the expected log-likelihood:
\begin{eqnarray}
	 \EV_{\calM\s{b}_i} \left[\log p(y_{1:\tau}|\calM\s{r}_j)\right] &\geq& {\calJ^{i,j}}({\calM\s{r}}, \boldsymbol\phi^{i,j}, \boldsymbol\eta) \quad \forall \boldsymbol\phi^{i,j}, \boldsymbol\eta
	\label{eqn:exp_log_hmm_wGMMvar_low_bound}
\end{eqnarray}
where we have defined:
\begin{eqnarray}
 \calJ^{i,j} (\calM\s{r}_j,\boldsymbol\phi^{i,j},\boldsymbol\eta) = \sum_{\beta_{1:\tau}} \pi\ssp{b}{i}{\beta}{1:\tau}
\sum_{\rho_{1:\tau}} \phi^{i,j}_{\rho_{1:\tau}|\beta_{1:\tau}} [ \log \pi\ssp{r}{j}{\rho}{1:\tau} + \sum_{t=1}^{\tau}  \calL(\calM\s{b}_{i,\beta_t} || \calM\s{r}_{j,\rho_t}) - \log \phi^{i,j}_{\rho_{1:\tau}|\beta_{1:\tau}}].
\label{eqn:exp_log_hmm_wGMMvar}
\end{eqnarray}
Maximizing the right hand side of \refeqn{eqn:exp_log_hmm_wGMMvar_low_bound}
with respect to $\boldsymbol\phi$ $\boldsymbol\eta$ finds the most accurate approximation to the real posterior within the restricted class, i.e., the one that achieves the tightest lower bound.

Finally, plugging \refeqn{eqn:exp_log_hmm_wGMMvar_low_bound} into \refeqn{eqn:cost_function}, we obtain a lower bound on the log-likelihood of the virtual sample:
\begin{eqnarray}
	{\calJ({\calM\s{r}})} &\geq& {\calJ}({\calM\s{r}}, \boldsymbol z, \boldsymbol\phi, \boldsymbol\eta) \quad \forall  \boldsymbol z, \boldsymbol\phi, \boldsymbol\eta
	\label{eqn:cost_all}
\end{eqnarray}
where we have defined:
\begin{eqnarray}
 {\calJ}({\calM\s{r}}, \boldsymbol z, \boldsymbol\phi, \boldsymbol\eta) &=& \sum_{i=1}^{K\s{b}} \sum_{j=1}^{K\s{r}} z_{ij} \log  \omega\s{r}_j - \sum_{i=1}^{K\s{b}} \sum_{j=1}^{K\s{r}} z_{ij} \log z_{ij} \nonumber\\
&+& \sum_{i=1}^{K\s{b}} \sum_{j=1}^{K\s{r}} z_{ij} N_i \sum_{\beta = 1}^{S} \pi\ssp{b}{i}{\beta}{1:\tau}
\sum_{\rho = 1}^{S} \phi^{i,j}_{\rho|\beta} \log \pi\ssp{r}{j}{\rho}{1:\tau}  \nonumber \\
&+& \sum_{i=1}^{K\s{b}} \sum_{j=1}^{K\s{r}} z_{ij} N_i \sum_{\beta = 1}^{S} \pi\ssp{b}{i}{\beta}{1:\tau}
\sum_{\rho = 1}^{S} \phi^{i,j}_{\rho_{1:\tau}|\beta_{1:\tau}} \sum_{t=1}^{\tau}  \calL(\calM\s{b}_{i,\beta_t} || \calM\s{r}_{j,\rho_t}) \nonumber \\
&-& \sum_{i=1}^{K\s{b}} \sum_{j=1}^{K\s{r}} z_{ij} N_i \sum_{\beta = 1}^{S} \pi\ssp{b}{i}{\beta}{1:\tau}
\sum_{\rho = 1}^{S} \phi^{i,j}_{\rho_{1:\tau}|\beta_{1:\tau}} \log \phi^{i,j}_{\rho_{1:\tau}|\beta_{1:\tau}}
\label{eqn:cost_all_terms}
\end{eqnarray}
To find the tightest possible lower bound to the log-likelihood of the virtual sample we need to solve
\begin{eqnarray}
	{\calJ({\calM\s{r}})} &\geq& \max_{ \boldsymbol z, \boldsymbol\phi, \boldsymbol\eta} {\calJ}({\calM\s{r}}, \boldsymbol z, \boldsymbol\phi, \boldsymbol\eta).
	\label{eqn:cost_all_max}
\end{eqnarray}
Starting from an initial guess for $\calM\s{r}$, the parameters can be estimated by maximizing \refeqn{eqn:cost_all_max} irteratively with respect to (\emph{E-step}) $\boldsymbol \eta$, $\boldsymbol\phi$, $\boldsymbol z$ and (\emph{M-step}) $\calM\s{r}$

\subsection{E-step}

The E-steps first considers the maximization of \refeqn{eqn:cost_all_max} with respect to $\boldsymbol \eta$ for fixed ${\calM\s{r}}$ $\boldsymbol z$ and $\boldsymbol\phi$, i.e.,
\begin{eqnarray}
	\hat{\boldsymbol\eta }=  \argmax_{ \boldsymbol\eta} {\calJ}({\calM\s{r}}, \boldsymbol z, \boldsymbol\phi, \boldsymbol\eta)
	\label{eqn:Estep_eta}
\end{eqnarray}
It can be easily verified that the maximization \refeqn{eqn:Estep_eta} does not depend on $\boldsymbol z$ and $\boldsymbol \phi$ can be carried out independently for each tuple $(i,j,\beta,\rho,m)$ using the result in Section \ref{sec:opt2} \cite{hershey2007approximating}, which gives:
\begin{eqnarray}
	\hat \eta^{(i,\beta),(j,\rho)}_{\ell|m} &=&  \frac{c^{(r),j}_{\rho,\ell} \exp \left\{ L_{G}(\calM^{(b),i}_{\beta,m}|| \calM^{(r),j}_{\rho,\ell})  \right\}}{\sum _{\ell'=1}^{M}c^{(r),j}_{\rho,\ell'} \exp \left\{ L_{G}(\calM^{(b),i}_{\beta,m}|| \calM^{(r),j}_{\rho',\ell}) \right\}}
	\label{eqn:eta_opt}
\end{eqnarray}
and that the terms in \refeqn{eqn:exp_ll_gmms} can then be computed for each $(i,j,\beta,\rho)$ as:
\begin{eqnarray}
	L(\calM\s{b}_{i,\beta} || \calM\s{r}_{j,\rho})= \sum_{m = 1}^{M} c^{(b),i}_{\beta,m} \log \sum_{\ell = 1}^{M} c^{(r),j}_{\rho,\ell} \exp \left\{ L_{G}(\calM^{(b),i}_{\beta,m}|| \calM^{(r),j}_{\rho,\ell})\right\}.
\end{eqnarray}

Next,  \refeqn{eqn:cost_all_max} is maximized with respect to $\boldsymbol \phi$ for fixed $({\calM\s{r}}$ $\boldsymbol z$ and $\boldsymbol\eta)$, i.e.,
\begin{eqnarray}
	\hat{\boldsymbol\phi }=  \argmax_{ \boldsymbol\phi} {\calJ}({\calM\s{r}}, \boldsymbol z, \boldsymbol\phi, \boldsymbol\eta)
	\label{eqn:Estep_phi}
\end{eqnarray}
The maximization does not depends on $\boldsymbol z$ and can be carried out independently for each pair $(i,j)$ with a backward recursion recursion \cite{hershey2008variational} that computes
	\begin{eqnarray}
	\hat\phi^{i,j}_t(\rho_{t-1},\rho_t,\beta_t) & = & \frac{\st{r}{j}{\rho}{t-1}{t} \exp \left\{ L(\calM\s{b}_{i,\beta_t} || \calM\s{r}_{j,\rho_t}) + \calL^{i,j}_{t+1}(\beta_{t},\rho_{t}) \right\}}{\sum_{\rho}^{} \st{r}{j}{\rho}{t-1}{} \exp \left\{ L(\calM\s{b}_{i,\beta_t} || \calM\s{r}_{j,\rho}) + \calL^{i,j}_{t+1}(\beta_{t},\rho) \right\}} \\
         \calL^{i,j}_t(\rho_{t-1},\beta_{t-1})            & = & \sum_{\beta =1}^{N} \st{b}{i}{\beta}{t-1}{} \log \sum_{\rho =1}^{N} \st{r}{j}{\rho}{t-1}{} \exp \left\{ L(\calM\s{b}_{i,\beta} || \calM\s{r}_{j,\rho})  + \calL^{i,j}_{t+1}(\beta,\rho) \right\}
	\end{eqnarray}
for $T = \tau,\dots,2$, where it is understood that $\calL^{i,j}_{\tau+1}(\beta_{t},\rho_{t}) = 0$, and terminates with
	\begin{eqnarray}
         \hat\phi^{i,j}_1(\rho_1,\beta_1) & = & \frac{\is{r}{j}{\rho}{1} \exp \left\{ L(\calM\s{b}_{i,\beta_1} || \calM\s{r}_{j,\rho_1}) + \calL^{i,j}_{2}(\beta_{1},\rho_{1}) \right\}}{\sum_{\rho}^{} \is{r}{j}{\rho}{{}} \exp \left\{ L(\calM\s{b}_{i,\beta_1} || \calM\s{r}_{j,\rho}) + \calL^{i,j}_{2}(\beta_{1},\rho) \right\}} \\
         \calJ^{i,j} (\calM\s{r}_j,\hat{\boldsymbol\phi}^{i,j},\boldsymbol\eta)           & = & \sum_{\beta =1}^{N} \is{b}{i}{\beta}{{}} \log \sum_{\rho =1}^{N} \is{r}{j}{\rho}{{}} \exp \left\{ L(\calM\s{b}_{i,\beta} || \calM\s{r}_{j,\rho}) + \calL^{i,j}_{2}(\beta,\rho) \right\}. \label{eqn:opt_el_hmm_1}
	\end{eqnarray}

Next, the maximization of \refeqn{eqn:cost_all_max} with respect to $\boldsymbol z$ for fixed ${\calM\s{r}}$ $\boldsymbol \phi$ and $\boldsymbol\eta$, i.e.,
\begin{eqnarray}
	\hat{\boldsymbol z }=  \argmax_{ \boldsymbol z} {\calJ}({\calM\s{r}}, \boldsymbol z, \boldsymbol\phi, \boldsymbol\eta)
	\label{eqn:Estep_z}
\end{eqnarray}
reduces to compute the $\hat{z}_{ij}$ as in \refeqn{eqn:zij} using \refeqn{eqn:opt_el_hmm_1} to approximate \refeqn{eqn:ExpLog}.

Finally, we compute the following summary statistics:
	\begin{eqnarray}
\nu_1^{i,j}(\sigma,\gamma) & = & \is{b}{i}{\gamma}{{}} \,\hat\phi^{i,j}_1(\sigma,\gamma)\\
\xi_{t}^{i,j}(\rho,\sigma,\gamma) & = & \left( \sum_{\beta=1}^{N}  \nu_{t-1}^{i,j}(\rho,\beta)  \, a^{(b),i}_{\beta,\gamma} \right)\, \hat\phi^{i,j}_{t}(\rho,\sigma,\gamma) \mbox{ for } t = 2,\dots,\tau \\
\nu_t^{i,j}(\sigma,\gamma) & = &   \sum_{\rho=1}^{N} \xi_t^{i,j}(\rho,\sigma,\gamma)
\mbox{ for } t = 2,\dots,\tau 
	\end{eqnarray}
and the aggregates 
	\begin{eqnarray}
&&\hat\nu_1^{i,j}(\sigma) \quad=\sum_{\gamma=1}^{N} \nu_1^{i,j}(\sigma,\gamma) \\
 &&\hat\nu^{i,j}(\sigma,\gamma) =\sum_{t=1}^{\tau} \nu_t^{i,j}(\sigma,\gamma) \\
&&\hat\xi^{i,j}(\rho,\sigma) = \sum_{t=2}^{\tau} \sum_{\gamma=1}^{N} \xi^{i,j}_t(\rho,\sigma,\gamma).
	\end{eqnarray}
	
The quantity $\nu_t^{i,j}(\sigma,\gamma)$ is the responsibility between state $\gamma $ of the HMM $\calM\s{b}_i$ and state $\sigma$ of the HMM $\calM\s{r}_j$ at time $t$, when modeling a sequence generated by $\calM\s{b}_i$.
Similarly, the quantity  $\xi^{i,j}_t(\rho,\sigma,\gamma)$ is the responsibility between a transition from state $\rho$ to state $\sigma$ (reached at time $t$) for the HMM $\calM\s{r}_j$  and state  $\gamma$ (at time $t$) of the HMM $\calM\s{b}_i$, when modeling a sequence generated by $\calM\s{b}_i$.
Consequently, the statistic $\hat\nu_1^{i,j}(\sigma)$ is the expected number of times that the HMM $\calM\s{r}_j$ starts from state $\sigma$, when modeling sequences generated by $\calM\s{b}_i$.
The quantity $\hat\nu^{i,j}(\sigma,\gamma)$ is the expected number of times that the HMM $\calM\s{b}_i$ is state $\gamma $ when the HMM $\calM\s{r}_j$ is in state $\sigma$, when both modeling sequences generated by $\calM\s{b}_i$.
Finally, the quantity  $\hat\xi^{i,j}(\rho,\sigma)$ is the expected number of transitions from state $\rho$ to state $\sigma$  of the HMM $\calM\s{r}_j$, when modeling sequences generated by $\calM\s{b}_i$.

\subsection{M-step}

The M-steps involves maximizing \refeqn{eqn:cost_all_max} with respect to ${\calM\s{r}} $ for fixed $\boldsymbol z$ $\boldsymbol\phi$ and $\boldsymbol \eta$, i.e.,
\begin{eqnarray}
	{{\hat\calM\s{r}} }=  \argmax_{ \calM\s{r}} {\calJ}({\calM\s{r}}, \boldsymbol z, \boldsymbol\phi, \boldsymbol\eta).
	\label{eqn:Mstep}
\end{eqnarray}
In the following, we detail the update rules for the parameters of the reduced model $\calM\s{r}$.

\subsubsection{HMMs mixture weights}
The re-estimation of the mixture weights, given the constraint $\sum_{j=1}^{K\s{r}} \omega\s{r}_j=1$, is solved using the result in Section \refeqn{sec:opt1}:
\begin{eqnarray}
	 {\omega\s{r}_j}^* &=& \frac{\sum_{i=1}^{K\s{b}} \hat z_{i,j}} {K\s{b}}.
	\label{eqn:est_omegaj}
\end{eqnarray}

\subsubsection{Initial state probabilities}

The cost function \refeqn{eqn:cost_all_max} factors independently for each $ \{\is{r}{j}{\sigma}{{}}\}_{\sigma = 1}^{N}$ ($j$ is fixed) and reduces to terms in the form:
\begin{eqnarray}
	{\calJ}({\calM\s{r}}, \boldsymbol z, \boldsymbol\phi, \boldsymbol\eta) & =&\sum_{\sigma=1}^{N}  \sum_{i=1}^{K\s{b}} \hat z_{i,j} N_i \hat\nu^{i,j}_1(\sigma) \,\, \log \pi^{(r),j}_{\sigma}.  
	\label{eqn:obj_f_pirj}
\end{eqnarray}
Considering the constraint $\sum_{\sigma=1}^{N} \is{r}{j}{\sigma}{{}} =1$, the results in Section \ref{sec:opt1} gives the update formulas
\begin{eqnarray}
	 {\is{r}{j}{\sigma}{{}}}^*
&=& \frac{\sum_{i=1}^{K\s{b}} \hat z_{i,j} N_i \sum_{\gamma=1}^{N} \hat\nu_1^{i,j}(\sigma)}{\sum_{\sigma'=1}^{N} \sum_{i=1}^{K\s{b}} \hat z_{i,j} N_i \sum_{\gamma=1}^{N} \hat\nu_1^{i,j}(\sigma')}.
	\label{eqn:est_pi_jrho}
\end{eqnarray}

\subsubsection{State transition probabilities}
Similarly, the cost function \refeqn{eqn:cost_all_max} factors independently for each $\{a^{(r),j}_{\rho,\sigma}\}_{\sigma=1}^{N}$ ($j$ and $\rho$ are fixed) and reduces to terms in the form:
\begin{eqnarray}
	{\calJ}(\{a^{(r),j}_{\rho,\sigma}\}_{\sigma=1}^{N}, \boldsymbol z, \boldsymbol\phi, \boldsymbol\eta) & =& \sum_{\sigma=1}^{N}  \sum_{i=1}^{K\s{b}} \hat z_{i,j} N_i  \hat\xi^{i,j}(\rho,\sigma) \,\, \log a^{(r),j}_{\rho,\sigma} 
	\label{eqn:obj_f_arj_rho}
\end{eqnarray}
Considering the constraint $\sum_{\sigma=1}^{N}  a^{(r),j}_{\rho,\sigma}=1$, the results in Section \ref{sec:opt1} gives the update formula
\begin{eqnarray}
	 {a^{(r),j}_{\rho,\sigma}}^* &=&  \frac{\sum_{i=1}^{K\s{b}} \hat z_{i,j} N_i \hat\xi^{i,j}(\rho,\sigma)}{\sum_{\sigma'=1}^{N} \sum_{i=1}^{K\s{b}} \hat z_{i,j} N_i \sum_{t=2}^{\tau} \hat\xi^{i,j}(\rho,\sigma') }
	\label{eqn:re_est_a}
\end{eqnarray}

\subsubsection{Eission probability density functions}
In general, in the cost function \refeqn{eqn:cost_all_max} factors independently for each $j,\rho,\ell$, and reduces to terms in the form:
\begin{eqnarray}
&&	{\calJ}(\calM^{(r),j}_{\rho,\ell}, \boldsymbol z, \boldsymbol\phi, \boldsymbol\eta)  = \nonumber\\ 
&&	
\sum_{i=1}^{K\s{b}}  \hat z_{i,j} N_i \sum_{\gamma=1}^{N}  \hat\nu^{i,j}(\rho,\gamma) \sum_{m = 1}^{M}  c^{(b),i}_{\beta,m} \,\,\hat \eta^{(i,\beta),(j,\rho)}_{\ell|m} \left( \log c^{(r),j}_{\rho,\ell} + L_{G}(\calM^{(b),i}_{\beta,m}|| \calM^{(r),j}_{\rho,\ell})  \right) 
	\label{eqn:obj_f_emit_rj_rho}
\end{eqnarray}
Using basic matrix calculus, and defining a weighted sum operator
	\begin{eqnarray}
\Omega_{j,\rho,\ell}( x(i,\beta,m) ) = \sum_i \hat z_{i,j} N_i \sum_{\beta} \hat\nu_t^{i,j}(\rho,\beta) \,\sum_{m = 1}^{M}  c^{(b),i}_{\beta,m} \, x(i,\beta,m)
	\end{eqnarray}
the parameters $\calM\s{r}$ are updated accordingly to:
	\begin{eqnarray}
{c^{(r),j}_{\rho,\ell} }^*    &=&  \frac{\Omega_{j,\rho,\ell}\left(\hat \eta^{(i,\beta),(j,\rho)}_{\ell|m} \right)}{\sum_{\ell'=1}^{M} \Omega_{j,\rho,\ell'}\left(\hat \eta^{(i,\beta),(j,\rho)}_{\ell'|m}\right) } \label{eq:update}\\
	{\mu^{(r),j}_{\rho,\ell} }^*    &=&  \frac{\Omega_{j,\rho,\ell}\left( \eta^{(i,\beta),(j,\rho)}_{\ell|m} \,\,\,\mu^{(b),i}_{\beta,m}\right)}{\Omega_{j,\rho,\ell}\left(\hat \eta^{(i,\beta),(j,\rho)}_{\ell|m} \right)}\\
	{\Sigma^{(r),j}_{\rho,\ell}}^* &=& \frac{\Omega_{j,\rho,\ell}\left(\hat \eta^{(i,\beta),(j,\rho)}_{\ell|m}\left[ \Sigma^{(b),i}_{\beta,m}  + (\mu^{(b),i}_{\beta,m}-\mu^{(r),j}_{\rho,\ell}) (\mu^{(b),i}_{\beta,m}-\mu^{(r),j}_{\rho,\ell})^t \right]\right)}{\Omega_{j,\rho,\ell}\left(\hat \eta^{(i,\beta),(j,\rho)}_{\ell|m}\right)}\label{eq:updateSigma}
	\end{eqnarray}
Equations (\ref{eq:update}-\ref{eq:updateSigma}) are all weighted averages over all base models, model states, and Gaussian components.

\section{Experiments}\label{sec:exp}

In this section, we present three empirical studies of the VHEM-H3M algorithm. Each application exploits some of the benefits of VHEM. First of all, instead of clustering HMMs on the parameter manifold, VHEM-H3M clusters HMMs directly through the distributions they represent. Given a collection of input HMMs, VHEM estimates a smaller mixture of novel HMMs that consistently models the distribution represented by the input HMMs. This is achieved by maximizing the log-likelihood of ``virtual'' samples generated from the input HMMs. As a result, the VHEM cluster centers are consistent with the underlying generative probabilistic framework. 

Second, VHEM allows to estimate models from large-scale data sets, by breaking the learning problem into smaller pieces. First, a data set is split into small non-overlapping portions and intermediate models are learned from each portion. Then, the final model is estimated from the intermediate models using the VHEM-H3M algorithm. While based on the same maximum-likelihood principles as direct EM estimation on the full data set, this VHEM estimation procedure has significantly lower memory requirements, since it is no longer required to store the entire data set during parameter estimation. In addition, since the intermediate models are estimated independently of each other, this estimation task can easily be parallelized. Lastly, the ``virtual'' samples (i.e., sequences) VHEM implicitly generates for maximum-likelihood estimation need not be of the same length as the actual input data for estimating the intermediate models. Making the virtual sequences relatively short will positively impact the run time of each VHEM iteration. This may be achieved without loss of modeling accuracy, as VHEM allows to compensate for shorter virtual training sequences by implicitly integrating over a virtually unlimited number of them.

\subsection{Hierarchical motion clustering}


\begin{figure}[h]
\begin{center}
\vspace{-.09in}
\includegraphics[width=0.85\textwidth]{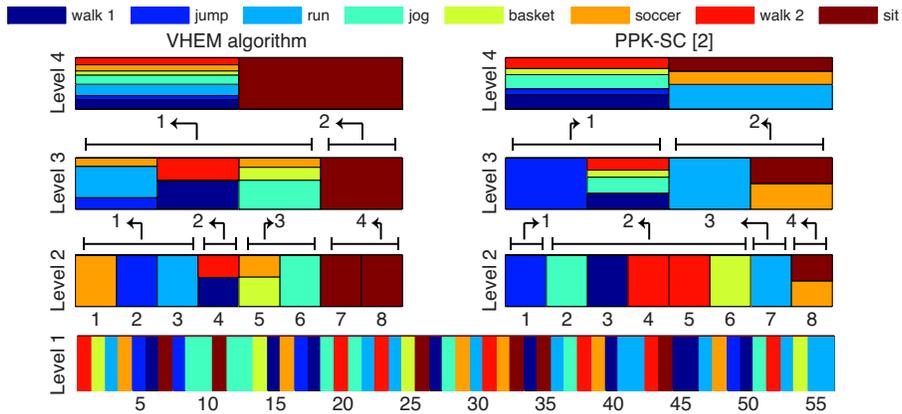}
\end{center}
\vspace{-.15in}
\caption{Hierarchical clustering of the MoCap dataset, with VHEM and SC-PPK.}
\label{fig:hier}
\end{figure} 



In this experiment we tested the VHEM algorithm on hierarchical motion clustering, using the Motion Capture dataset (http://mocap.cs.cmu.edu/), which is a collection of time-series data representing human motions.
In particular, we start from $K_1 = 56$ motion examples form $8$ different classes, and learn a HMM for each of them, forming the first level of the hierarchy.
A tree-structure is formed by successively clustering HMMs with the VHEM algorithm, and using the learned cluster centers as the representative HMMs at the new level.
The second, third and fourth levels of the hierarchy correspond to, respectively, $K_2 = 8$, $K_3 = 4$ and $K_4 = 2$.

The hierarchical clustering obtained with VHEM is illustrated in Figure \ref{fig:hier} (left).
In the first level, each vertical bar represents a motion sequence, with different colors indicating different ground-truth classes.
In the second level, the $8$ HMM clusters are shown with vertical bars, with the colors indicating the proportions of the motion classes in the cluster.
Almost all clusters are populated by examples from a single motion class (e.g., ``run'', ``jog'', ``jump''), which demonstrates that VHEM can group similar motions together. 
We note an error of the VHEM in clustering a portion of the ``soccer'' examples with  ``basket''. 
Moving up the hierarchy, the VHEM algorithm clusters similar motions classes together (as indicated by the arrows), and at the last (Level 4) it creates a dichotomy between the ``sit'' and the rest of the motion classes. This is a desirable behavior  as the a the kinetics of the ``sit'' sequences 
(i.e., sitting on a stool and going down) 
are considerably different form the rest.
On the right of Figure \ref{fig:hier}, the same experiment is repeated using spectral clustering in tandem with PPK similarity (SC-PPK) \cite{jebara2007}. The SC-PPK clusters motions sequences properly, however it incorrectly aggregates the ``sit'' and ``soccer'', and produces a last level  (Level 4) not well interpretable.

While VHEM has lower Rand-index than SC-PPK at Level 2 ($0.940$ vs. $0.973$), it has higher Rand-index at Level 3 ($0.775$ vs. $0.737$) and Level 4 ($0.591 $ vs. $0.568 $). This suggests that the novel HMMs cluster centers learned by VHEM retain more information that the spectral cluster centers.

{\small
\begin{minipage}[b]{.60\textwidth}
	\begin{tabular}{ccccccc}
		& \multicolumn{3}{c}{\bf annotation}  &\multicolumn{3}{c}{\bf retrieval} 
		\\
			& P & R & F & MAP & AROC & P@10 \\
			\hline 
			\\
		HEM-H3M 	& 0.470	&0.210	&0.258	&0.438	&0.700	&0.450\\
		EM-H3M	&0.415	&0.214	&0.248	&0.423	&0.704	&0.422\\
		HEM-DTM	&0.430	&0.202	&0.252	&0.439	&0.701	&0.453\\
                \end{tabular}
	\captionof{table}{Annotation and retrieval performance on CAL500, for VHEM-H3M, EM-H3M and HEM-DTM\cite{coviello2010}}
	  \label{tab:cal500}
\end{minipage}\qquad \hspace{0.01in}
\begin{minipage}[b]{.35\textwidth}	
\begin{tabular}{lcc}
& \multicolumn{2}{c}{{\bf classification}}\\
		& VHEM-H3M	& EM-H3M  \\
				\hline 
			\\
	 $\tau = 5$  & 0.569 & 0.349  \\
	  $\tau =10$  	& 0.570 & 0.389 \\
	 $\tau =15$ & 0.573 & 0.343\\
	 \end{tabular}
         \captionof{table}{Online hand-writing classification accuracy (20 characters)}
  \label{tab:letter}
\end{minipage}     
}
\vspace{-.15in}
\subsection{Automatic music tagging}

In this experiment we evaluated VHEM-H3M on automatic music tagging.
We considered the CAL500 collection form Barrington et al. \cite{music:turnbull08}, which consists in 502 songs 
and provides binary annotations with respect to a vocabulary ${\cal V}$ of 149 tags, ranging from genre and instrumentation, to mood and usage. 
To represent the acoustic content of a song we extract a time series of audio features ${\cal Y} = \{y_1,\dots,y_T\}$, by computing  the first 13 Mel frequency cepstral coefficients (MFCC) \cite{Rabiner93} over half-overlapping windows of $92$ms of audio signal, augmented with first and second derivatives. 

Automatic music tagging is formulated as a supervised multi-class labeling problem \cite{carneiro07},
where each class is a tag from ${\cal V}$. We model tags with H3M probability distributions over the space of audio fragments (e.g., sequences of $\tau = 125$ audio features, which approximately corresponds to 6 seconds of audio). 
Each tag model is learned from audio-fragments extracted from relevant songs in the database, using the VHEM-H3M.
The database is first processed at the song level, using the EM algorithm to learn a H3M for each song from a dense sampling of audio fragments.
For each tag, the song-level H3Ms that are relevant to the tag are pooled together to form a big H3M, and the VHEM algorithm is used to learn the final tag-model.

In table \ref{tab:cal500} we present a comparison of the VHEM-H3M algorithm with the standard EM-H3M algorithm and a state-of-the-art auto-tagger (HEM-DTM) \cite{coviello2010}, which uses the dynamic texture mixture model and an efficient HEM algorithm,
on both annotation an retrieval on the CAL500 dataset. Annotation is measured with precision (P), recall (R), f-score (F), and retrieval is measured with mean average precision (MAP), area under the operating characteristic curve (AROC), and precision at the first 10 retrieved objects (P@10). 
All reported metrics are averages over the 98 tags that have at least 30 examples in CAL500, and are result of 5 fold-cross validation.
VHEM-H3M  achieves better performance over EM-H3M (except on precision and AROC which are comparable) and strongly improves the top of the ranking list, as demonstrated by the higher P@10 score.
Performance of VHEM-H3M and HEM-DTM are close on all metrics with only slight variations, except on annotation precision where VHEM-H3M registers a significantly higher score.%


\subsection{Online hand-writing recognition}

In this experiment we investigated the performance of the VHEM-H3M algorithm on classification of on-line hand-writing.
We considered the Character Trajectories Data Set \cite{williams2006extracting}, which consists in $2858$ examples of characters 
from the same writer, and used half of the data for training and half for testing.
An HMM (with $N=2$ and $M=1$) was first learned from each of the training sequences using the EM algorithm. For each letter, all the relevant HMMs were clustered with the VHEM to form a H3M with $K\s{r} = 2$ components.
We repeated the same experiment using the EM-H3M algorithm
 directly on all the relevant sequences in the train data. For each letter, we allowed the EM algorithm up to three times the total running time of the VHEM (including the estimation of the corresponding intermediate HMMs).
Table \ref{tab:letter} lists classification accuracy on the test set, for VHEM-H3M, using different values of $\tau$, and for the corresponding runs of EM-H3M.
A small $\tau$ suffices to provide a regular estimate, and simultaneously determines shorter running times for VHEM (under $2$ minutes for all $20$ letters).
On the other hand, the EM algorithm needs to evaluate the likelihood of all the original sequences at each iteration, which determines slower iterations, and prevents the EM from converging to effective estimates in the time allowed. 

\section{Conclusion}
In this paper, we present a variational HEM (VHEM) algorithm for clustering HMMs through their distributions. Moreover, VHEM summarizes each cluster by estimating a new HMM as cluster center.
We demonstrate the efficacy of this algorithm for various applications, including hierarchical motion clustering, semantic music annotation, and online hand-writing recognition.
%

\section{Appendix on useful optimization problems}

\subsection{}\label{sec:opt1}
The optimization problem
\begin{eqnarray}
	\max_{\alpha_\ell} & & \sum_{\ell=1}^{L} \beta_\ell \log \alpha_\ell    	\label{eqn:emax_form}\\
	\mbox{s.t.}			& & \sum_{\ell=1}^{L} \alpha_\ell = 1  \nonumber\\
					& & \alpha_\ell \geq 0,\, \forall \ell \nonumber
\end{eqnarray}
is optimized by 
\begin{eqnarray}
	\alpha_\ell^* = \frac{\beta_\ell}{\sum_{\ell'=1}^{L} \beta_\ell'}.
	\label{eq:op1}
\end{eqnarray}
This can be easily computed with the optimization
\begin{eqnarray}
	\{\alpha_\ell^*\} &=& \argmax_{\alpha_\ell} \sum_{\ell=1}^{L} \beta_\ell \log \alpha_\ell + \lambda \left( \sum_{\ell=1}^{L} \alpha_\ell - 1 \right) \nonumber 
\end{eqnarray}
where the second term is a Lagrangian term for the weights to sum to $1$, and noticing that the positivity constraints are automatically satisfied by \refeqn{eq:op1}.

\subsection{}\label{sec:opt2}
The optimization problem
\begin{eqnarray}
	\max_{\alpha_\ell} & & \sum_{\ell=1}^{L} \alpha_\ell \left( \beta_\ell - \log \alpha_\ell  \right) \label{eqn:gmax_form} \\
	\mbox{s.t.}			& & \sum_{\ell=1}^{L} \alpha_\ell = 1  \nonumber\\
					& & \alpha_\ell \geq 0,\, \forall \ell \nonumber
\end{eqnarray}
is optimized by 
\begin{eqnarray}
	\alpha_\ell^* = \frac{\exp \beta_\ell}{\sum_{\ell'=1}^{L} \exp  \beta_\ell'}.
	\label{eq:op2}
\end{eqnarray}
This can be easilly computed with the optimization
\begin{eqnarray}
	\{\alpha_\ell^*\} &=& \argmax_{\alpha_\ell} \sum_{\ell=1}^{L} \alpha_\ell \left( \beta_\ell - \log \alpha_\ell  \right) + \lambda \left( \sum_{\ell=1}^{L} \alpha_\ell - 1 \right) \nonumber 
\end{eqnarray}
where the second term is a Lagrangian term for the weights to sum to $1$,and noticing that the positivity constraints are automatically satisfied by \refeqn{eq:op2}.

\vspace{-3mm}
\fontsize{10}{10}
\bibliographystyle{./IEEEtran}
\bibliography{References}

\end{document}